%% file: main.tex
\documentclass[acmtog]{acmart}
\AtBeginDocument{%
  }

\setcopyright{acmlicensed}
\copyrightyear{2025}
\acmYear{2025}
\acmDOI{3757377.3763858}
\acmConference[SA Conference Papers '25]{}{December 15--18,
  2025}{Hong Kong, Hong Kong}
\acmISBN{979-8-4007-2137-3/25/12}
\usepackage{xcolor}
\usepackage{graphicx}
\usepackage{subcaption}
\definecolor{pink}{RGB}{255,0,0}

\begin{document}

\title{Clustered Error Correction with Grouped 4D Gaussian Splatting}

\author{Taeho Kang}
\email{taeho.kang@hcs.snu.ac.kr}
\orcid{0000-0002-4556-5588}
\affiliation{%
  \institution{Seoul National University}
  \city{Seoul}
  \country{South Korea}
}

\author{Jaeyeon Park}
\email{jypark@snu.ac.kr}
\orcid{0009-0002-0952-7573}
\affiliation{%
  \institution{Seoul National University}
  \city{Seoul}
  \country{South Korea}
}

\author{Kyungjin Lee}
\email{kjlee818@gmail.com}
\orcid{0000-0002-7120-9595}
\affiliation{%
  \institution{Seoul National University}
  \city{Seoul}
  \country{South Korea}
}

\author{Youngki Lee}
\email{youngki.lee@gmail.com}
\orcid{0000-0002-1319-7071}
\affiliation{%
  \institution{Seoul National University}
  \city{Seoul}
  \country{South Korea}
}
\newcommand{\methodname}{our method}

\renewcommand{\shortauthors}{Kang et al.}



\input{Sections/00_abstract}

\maketitle
\input{Sections/01_intro}
\input{Sections/02_related_works}
\input{Sections/03_background}
\input{Sections/04_method}
\input{Sections/05_evaluation}
\input{Sections/06_conclusion}
\begin{acks}
This work was supported by the National Research Foundation of Korea(NRF) grant funded by the Korea government(MSIT) (No. RS-2024-00463802, No. RS-2023-00218601).
\end{acks}


\include{Sections/99_supplementary}
\bibliographystyle{ACM-Reference-Format}
\bibliography{main}

\end{document}

%% file: Sections/00_abstract.tex
\begin{abstract}
Existing 4D Gaussian Splatting (4DGS) methods struggle to accurately reconstruct dynamic scenes, often failing to resolve ambiguous pixel correspondences and inadequate densification in dynamic regions. We address these issues by introducing a novel method composed of two key components: (1) Elliptical Error Clustering and Error Correcting Splat Addition that pinpoints dynamic areas to improve and initialize fitting splats, and (2) Grouped 4D Gaussian Splatting that improves consistency of mapping between splats and represented dynamic objects. Specifically, we classify rendering errors into missing-color and occlusion types, then apply targeted corrections via backprojection or foreground splitting guided by cross-view color consistency. Evaluations on Neural 3D Video and Technicolor datasets demonstrate that our approach significantly improves temporal consistency and achieves state-of-the-art perceptual rendering quality, improving 0.39dB of PSNR on the Technicolor Light Field dataset. Our visualization shows improved alignment between splats and dynamic objects, and the error correction method's capability to identify errors and properly initialize new splats. Our implementation details and source code are available at https://github.com/tho-kn/cem-4dgs.
\end{abstract}

%% file: Sections/01_intro.tex
\section{Introduction} 
\label{introduction}
As augmented and virtual reality gain more attention, immersive content creation becomes more important. One way to provide hyperrealistic content is to incorporate the real world into computer graphics. Traditional 3D representations, such as meshes or voxel grids, are effective for handcrafted 3D models; however, they have limitations for reconstructing complex real-world scenes from images. In recent years, NVS~(Novel View Synthesis) has become a popular method for the hyperrealistic representation of scenes. NeRF~(Neural Radiance Field)~\cite{mildenhall2020nerf} showed promising results in synthesizing views in extremely high quality, given camera transformation in the scene with neural representation. Later efforts improved its efficiency~\cite{chen2022tensorf, M_ller_2022} and rendering quality~\cite{barron2021mipnerf}. 3DGS~(3D Gaussian Splatting)~\cite{kerbl3Dgaussians} has gained attention due to its efficiency, allowing real-time rendering. The original 3DGS is limited to synthesizing novel views for a static scene. Many efforts expand the 3DGS methods into 4DGS for dynamic scenes prevalent in the real world~\cite{wu20234d,yang2023real4dgs,guo2024motionaware,li2023spacetime, lee2024ex4dgs, bae2024ed3dgs}.

\begin{figure}[t] \centering \includegraphics[width=0.9\linewidth]{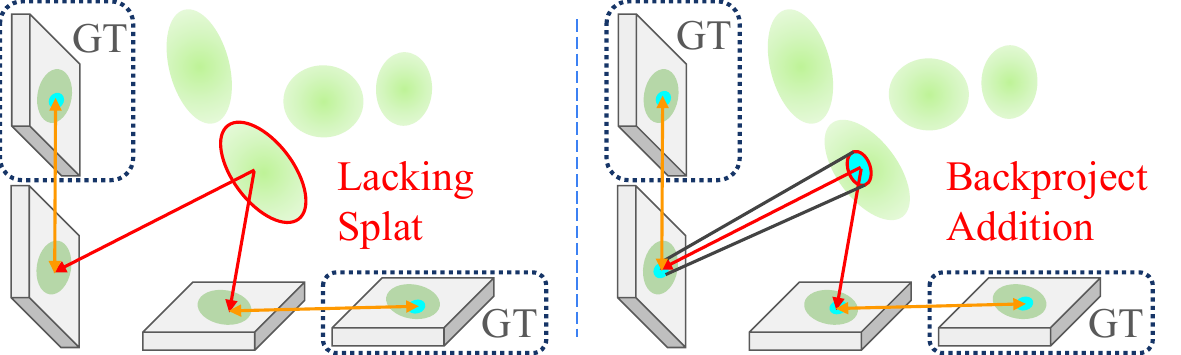} \caption{Failure of densification results in a lack of splat for the blue colored area on the green surface. By backprojecting the elliptical error cluster's ground truth color, we effectively add a Gaussian splat that corrects pixel error.} \label{fig:lacking_splat} \centering \includegraphics[width=0.9\linewidth]{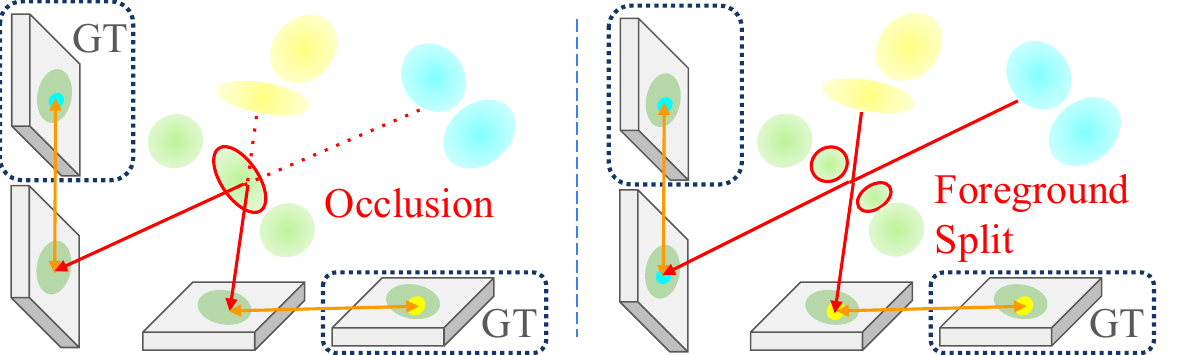} \caption{Due to occluding splat, there is a green visual error where ground truth colors should be yellow and blue, respectively. Splitting and optimizing the occluding foreground splat to cover only the correct area makes occluded splats with ground-truth colors visible from both views.} \label{fig:occlusion} \end{figure}

\begin{figure}[t]
    \centering
    \includegraphics[width=\linewidth]{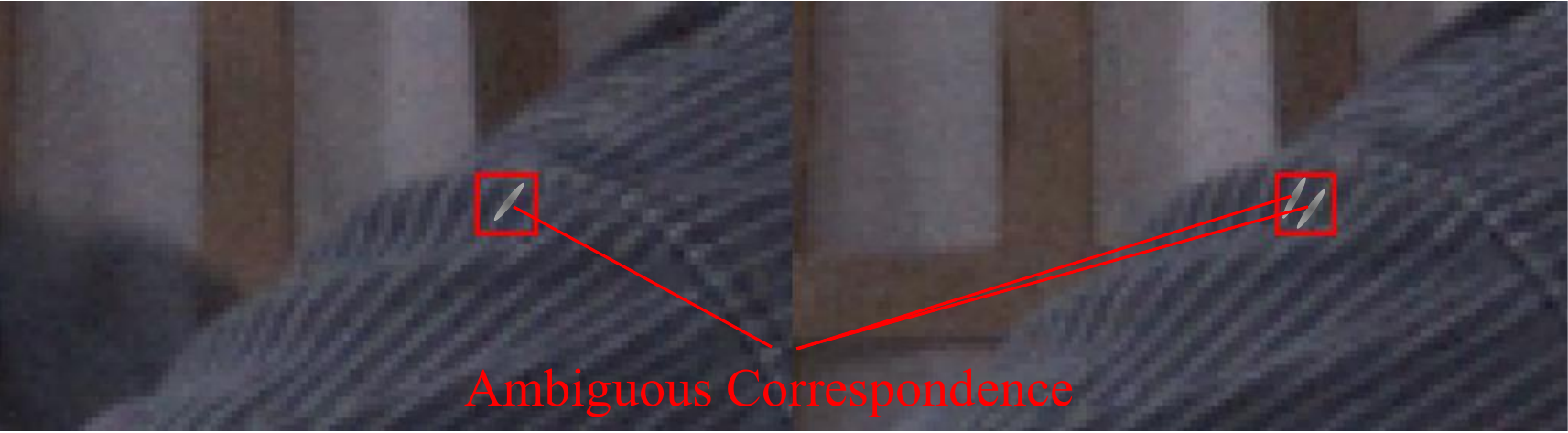}
    \caption{Looking at the image as a whole, the correspondence between stripes is apparent between different frames. However, focusing on a small area like splats, the correspondence between them becomes ambiguous, as shown in the red box in the same coordinate. A splat to represent part of a stripe can be mapped to its neighboring stripe with a dynamic transform during the optimization process.}
    \label{fig:correspondence}
\end{figure}

Despite the effort, several challenges remain for dynamic scene NVS~(Novel View Synthesis). The densification strategy is not practical for dynamic areas in 4D Gaussian Splatting. In particular, dynamic details are not effectively reconstructed because of the lack of splats. 3D Gaussian Splatting uses the average viewpoint gradient to determine which splats to densify. However, the extension to a 4D scene causes the appearance of the splat to vary over time, and the vanilla approach does not pinpoint the dynamic area to densify, as it is significantly impacted by temporal transform error and visibility changes. Furthermore, optimizing time-variant splats for details of a dynamic object is not trivial. Each splat covers a small area with one color, and there are multiple possible correspondences with similarly colored pixels near each other in adjacent frames, as shown in Fig.~\ref{fig:correspondence}. Extending the dynamic splat temporally with the correct motion is challenging with this ambiguity. Previous efforts, such as Embedding-Based Deformable 3D Gaussian Splatting (E-D3DGS)~\cite{lee2024ex4dgs} and Spacetime Gaussians (STG)~\cite{li2023spacetime}, have attempted to mitigate these issues by adapting multiview DSSIM loss for densification or uniformly sampling erroneous patches. However, these strategies lack precise localization of error regions and explicit modeling of splat correspondences across frames, which are crucial for detailed and temporally stable reconstruction. Consequently, dynamic scenes rendered by these methods still exhibit noticeable flickering and compromised detail reconstruction, particularly in highly dynamic areas or around fine textures.

To explicitly address these limitations, we propose a novel method that consists of two complementary components: (1) an elliptical clustering method that accurately identifies and localizes high-error pixel regions for precise injection of Gaussian splats, and (2) a group-based dynamic splat modeling approach that ensures temporal consistency by enforcing shared motion transforms across grouped splats. Our elliptical clustering enables accurate pinpointing of regions requiring correction, effectively reducing redundant splats and enhancing visual accuracy. Simultaneously, our group-based modeling approach resolves the ambiguity of splat correspondences, significantly reducing temporal jitter by ensuring consistent splat tracking across frames (see Fig.~\ref{fig:qualitative_consistency}).

Specifically, we first detect and cluster rendering errors into elliptical shapes by comparing the rendered frames with ground-truth reference views. Each error cluster is further categorized into missing-color or occlusion errors by performing cross-view color consistency analysis. For missing-color errors, new splats are injected through targeted backprojection; occlusion errors are corrected by foreground splat splitting to resolve depth inaccuracies. These targeted correction strategies substantially improve spatial precision and perceptual quality, rendering dynamic scene details more faithfully.

Experiments on widely used benchmarks, including Neural 3D Video~\cite{neural3dvideo} and Technicolor~\cite{Sabater2017technicolor} datasets, demonstrate that our proposed method achieves state-of-the-art perceptual quality, significantly outperforming existing approaches in metrics such as PSNR, DSSIM, and LPIPS. Furthermore, our temporal modeling consistently enhances temporal stability and reduces visual artifacts, clearly advancing dynamic scene rendering. We will publicly release our implementation and detailed source code to support future research and applications. Our main contributions are the following:

\begin{itemize}
\item{We introduce a pixel error clustering approach that accurately identifies the elliptical region to correct with splats.}
\item{We propose a dynamic splats grouping that clarifies the correspondence between dynamic areas across different frames.}
\item{We identify two types of errors in the details of 4DGS unresolved by densification and devise a multi-view driven method to pinpoint coordinates to introduce new splats.}
\item{Our method achieves state-of-the-art visual quality in Neural 3D Video and Technicolor datasets.}
\end{itemize}

%% file: Sections/02_related_works.tex
\section{Related Works}
\subsection{Novel View Synthesis for Static Scenes}
Traditional approaches solve for precise geometries~\cite{visualhulls, sfmrevisited}, the novel view rendering of meshes or point clouds generated by MVS has limited accuracy. For high-quality NVS, image-based rendering, given dense image samples, was a popular approach.

NeRF~\cite{mildenhall2020nerf} is the pioneering work that initiated this trend, using a multilayer perceptron (MLP) to learn the radiance field and volumetric rendering to synthesize images from any viewpoint.
Subsequent work has aimed to boost the quality and efficiency of differentiable-volume rendering. One direction focuses on improving the sampling strategy to reduce the number of point queries~\cite{piala2021terminerf, Neff_2021, M_ller_2022}. Another group incorporates explicit and localized neural representations to enhance model compactness or reduce rendering time \cite{barron2023zipnerf, chen2022tensorf, yu2021plenoxels, hu2023trimiprf, liu2021neural,  reiser2021kilonerf, Suhail_2022_CVPR, yu2021plenoctrees}.
Using image-based priors or depth information has proven effective in NeRF-based models. For example, Roessle et al. \cite{roessle2022dense} and NerfingMVS \cite{wei2021nerfingmvs} use a pre-trained network to estimate the dense depth from the sparse SfM points and supervise the NeRF model with the estimated depth.
MonoPatchNeRF\cite{wu2024monopatchnerf} integrates patch-based monocular priors for more precise handling of depth and normal maps. 

As another significant stream of novel view synthesis, 3D Gaussian Splatting~\cite{kerbl3Dgaussians} (3DGS) represents a scene using a 3D Gaussian collection. 3D coordinates, color, size, density, orientation, and spherical harmonics characterize each Gaussian in 3DGS. These parameters are optimized using an efficient differentiable rasterization algorithm, starting from a set of Structure-from-Motion (SfM) points to represent a given 3D scene best.

\subsection{Novel View Synthesis for Dynamic Scenes}
Synthesizing dynamic scenes introduces additional complexities, notably the need to capture temporal coherence across frames accurately. Dynamic extensions of NeRF~\cite{mildenhall2020nerf} (e.g., Nerfies~\cite{park2021nerfies}, HyperNeRF~\cite{park2021hypernerf}, D-NeRF~\cite{pumarola2021d}) use deformation networks and latent space embeddings to represent time-varying geometry and appearance, or combines NeRF with time-conditioned latent codes to represent dynamic scenes.~\cite{li2021neural, li2022neuraldynerf}. More recent approaches like K-Planes~\cite{fridovichkeil2023kplanes}, HexPlane~\cite{cao2023hexplane}, and Tensor4D~\cite{shao2023tensor4d} factorize the 4D spacetime domain into 2D feature planes for a more compact model size.

Efforts have also been made to extend 3D Gaussian Splatting (3DGS) into dynamic scenes \cite{wu20234d, yang2023real4dgs}.
Some approaches focus on multi-view images from the dynamic monocular or stereo camera~\cite{huang2024scgs, lu2024gags, liang2025himormonoculardeformablegaussian, lin2024gaussianflow}, such as the D-NeRF~\cite{pumarola2021d} and HyperNeRF~\cite{park2021hypernerf} datasets. SC-GS~\cite{huang2024scgs} and HiMoR~\cite{liang2025himormonoculardeformablegaussian} use interpolation of multiple control points to represent dynamicity. Others focus on compressing the 4D Gaussian Splatting representation for storage efficiency~\cite{Lee_2024_CVPR}.
Multi-camera setups have been widely explored. Early approach fixes the Gaussians' number and properties and optimizes their position and orientation~\cite{luiten2023dynamic}. 
4DGaussians~\cite{wu20234d} and D3DGS~\cite{yang2023deformable} reconstruct scenes using 3D Gaussians with a deformation field, while E-D3DGS \cite{bae2024pergaussian} further defines deformations as functions of per-Gaussian and temporal embeddings. ST-4DGS~\cite{Li2024ST-4DGS} enhances the deformation-based approach by incorporating temporal shape regularization and temporal-aware density control. The other direction is to use a Gaussian distribution in the 4D space and slice it in the time axis to get the 3D distribution~\cite{duan:2024:4drotorgs, yang2023real4dgs}.
STG~\cite{li2023spacetime} proposed modeling dynamicity with temporal opacity, parametric motion/rotation, and splatted feature rendering. 
SWinGS~\cite{shaw2024swingsslidingwindowsdynamic} uses sliding-window optimization with adaptively sampled windows.
Ex4DGS~\cite{lee2024ex4dgs} explicitly extracts dynamic Gaussian splats and models its motion with a keyframe interpolation-based dynamic transform. Concurrent work uses sparse 4D grid anchors~\cite{cho20244dscaffoldgaussiansplatting} and disentangles temporal components from a 4D Gaussian distribution~\cite{feng2025disentangled4dgaussiansplatting}.

Some prior works~\cite{luiten2023dynamic, Li2024ST-4DGS} use local rigidity loss for effective temporal modeling. In our experiment with similar loss, the higher weights degraded the visual quality, while the lower weights did not produce a consistent object-aligned motion. Our method improves the quality of 4D Gaussian Splatting by modeling more accurate correspondences of dynamic objects across frames using shared dynamic properties for splats and splitting the group carefully.

\subsection{Gaussian Splatting Densification Strategies}
Densification in 3DGS refers to the process of adding Gaussians to enhance the scene representation. One of the challenges with 3DGS is that it can excessively increase the number of Gaussians without explicitly constraining the real geometric structure, leading to numerous redundant Gaussians and significant memory consumption. 

Recent methods aim to optimize densification in 3DGS by reducing redundancy. For instance, some approaches transform 3D Gaussians into a more compact format by minimizing the number of Gaussians without compromising visual quality \cite{fan2024lightgaussian, morgenstern2024compact}. Another approach employs a progressive propagation strategy to guide the densification of 3D Gaussians, leveraging priors from existing reconstructed geometries and patch-matching techniques to produce new Gaussians \cite{cheng2024gaussianpro}. Some approaches consider pixel error for densification. One accumulates pixel errors that correspond to splats instead of viewpoint gradient~\cite{revisingdensification}. Pixel-GS rescales the gradient used for densification using the number of pixels each splat covers~\cite{pixelgs}. However, these methods are restricted to static scenes and do not apply to dynamic objects. Our proposed method complements the limitations of naive 3DGS densification for the 4D Gaussian Splatting setup, and improves the temporal fidelity of dynamic 3DGS methods while maintaining efficiency.

For dynamic scenes, most methods use densification similar to 3DGS with a gradient from L1 and DSSIM loss~\cite{wu20234d, li2023spacetime, yang2023real4dgs, lee2024ex4dgs}. STG~\cite{li2023spacetime} additionally samples new Gaussians along the rays of patches of pixels with large errors to improve faraway static object quality. E-D3DGS~\cite{bae2024pergaussian} periodically adds a multi-view DSSIM loss to induce Gaussian densification. They only indirectly determine where more Gaussian splats are needed to improve quality with simple approaches. Our method pinpoints where and what Gaussian splats should be added to correct the error in the rendered frame, and expands it temporally following the existing splat group's motion.

%% file: Sections/03_background.tex
\section{Preliminary}
\label{sec:3dgs}
\subsection{3D Gaussian Splatting}
3D Gaussian splatting represents scenes using sparse 3D Gaussian distributions, allowing for efficient and high-quality rendering of static 3D scenes. Each Gaussian is defined by a covariance matrix \(\Sigma \in \mathbb{R}^{3 \times 3}\) and a mean \(\mu \in \mathbb{R}^{3}\). The Gaussian function is given by:

\begin{equation}
\mathbf{G}(\mathbf{x}) = e^{-\frac{1}{2} (\mathbf{x} - \mu)^{\top} \Sigma^{-1} (\mathbf{x} - \mu)}, 
\quad where \quad
\Sigma = R S S^{\top} R^{\top}.
\end{equation}

To ensure the positive semi-definiteness of \(\Sigma\) and simplify the learning process, \(\Sigma\) is decomposed into a scaling matrix \(S\) and a rotation matrix \(R\). Rendering from a specific viewpoint involves transforming \(\Sigma\) into the camera coordinates using the viewing transformation matrix \(W\) and the Jacobian \(J\) of the affine approximation of the projective transformation:
\begin{equation}
\Sigma' = J W \Sigma W^{\top} J^{\top}.
\end{equation}

Each 3D Gaussian is parameterized by position \(\mathbf{x} \in \mathbb{R}^{3}\), color defined by spherical harmonics coefficients \(\mathbf{c} \in \mathbb{R}^{k}\) (where \(k\) means the number of SH coefficients), rotation \(\mathbf{r} \in \mathbb{R}^{4}\) represented by quaternion, scale \(\mathbf{s} \in \mathbb{R}^{3}\), and opacity \(o \in \mathbb{R}\). The color \(C\) of each pixel(\(\textbf{p}\)) is determined using point-based alpha blending, influenced by the Gaussians overlapping that pixel:

\begin{equation}
\begin{aligned}
\mathbf{C}(\mathbf{p}) &= \sum_{i=1}^{N} \mathbf{T}_i \alpha_i \mathbf{c}_i, \quad \text{where} \quad
\alpha_i = \sigma_i \mathbf{G}_i(\mathbf{x}),\ \mathbf{T}_i = \prod_{j=1}^{i-1} (1 - \alpha_j)
\end{aligned}
\end{equation}

where \(\alpha_i\) represents the density of each Gaussian, \(\textbf{T}_i\) is the transmittence, and \(\mathbf{c}_i\) is the color of gaussians overlapped with the pixel.

\subsection{Explicit 4D Gaussian Splatting}
\label{sec:ex4dgs}
We choose Fully Explicit 4D Gaussian Splatting~\cite{lee2024ex4dgs} as a baseline, explicitly distinguishing scenes between static and dynamic splats.

Static splats are modeled with a simple linear displacement \(\mathbf{d} \in \mathbb{R}^{3}\) in addition to the vanilla 3D Gaussian Splats. The position of the static splat at time $t$ is:
\begin{equation}
\mathbf{x_s}(t) = \mathbf{x_s}(0) + t*\mathbf{d}
\end{equation}
After optimization, splats for the dynamic areas have high displacement. They are periodically extracted and promoted to dynamic splats.

Dynamic splats have four dynamic properties in addition to 3DGS properties: position \(\mathbf{x_d} \in \mathbb{R}^{3 \times K}\), rotation \(\mathbf{r_d} \in \mathbb{R}^{4 \times K}\) replacing static counterparts, and time-variant opacity weight center \(\mathbf{o_c} \in \mathbb{R}^{2}\) and variance \(\mathbf{o_v} \in \mathbb{R}^{2}\), where $K$ is the number of keyframes. The position and rotation are explicitly given as values for the keyframes every 10 frames, and the cubic Hermite spline (CHip) and slerp~\cite{slerp} interpolation are used for position and rotation in the intermediate frames. Dynamic splat's opacity is a multiplication of opacity $o$ and opacity weight $o_w(t)$ parameterized by \(\mathbf{o_c}\) and \(\mathbf{o_v}\). 
The piecewise function, composed of two Gaussian segments and a constant segment, models the temporal behavior of a dynamic object: the Gaussian segments capture its appearance and disappearance with individual centers and variances. In contrast, the constant segment between Gaussian segments represents its stable existence. The supplementary material provides further details of this formulation.

%% file: Sections/04_method.tex
\section{Method}

\begin{figure*}[h]
    \centering
    \includegraphics[width=0.9\linewidth]{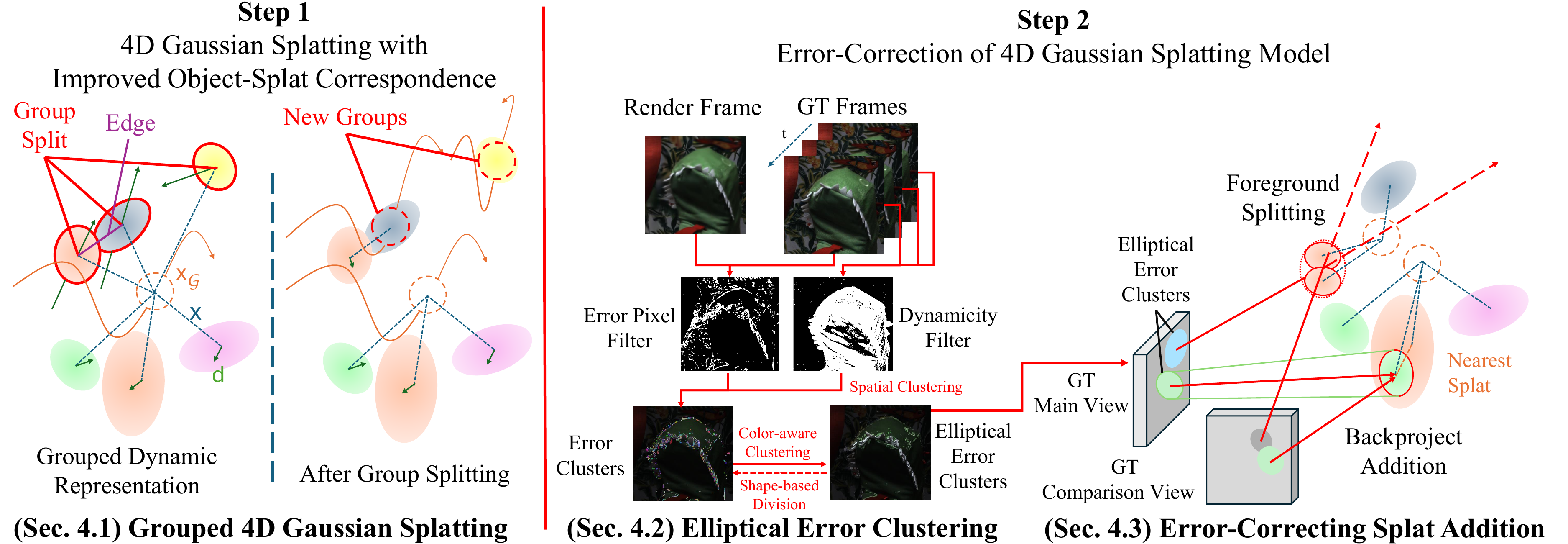}
    \caption{Overall framework of our method. We aim to improve the rendering quality by introducing dynamic splats that correct observed pixel errors in the training set. Grouped 4DGS uses shared dynamic transforms for a group of dynamic splats to improve temporally consistent correspondence between splats and dynamic scene objects. (Sec.~\ref{sec:dynamic_group}) Elliptical Error Clustering extracts error pixels and clusters pixels into an elliptical shape, ready for backprojection to initialize a Gaussian splat. (Sec.~\ref{sec:candidate_clustering}) Error-Correcting Splat Addition corrects the pixel error of clusters caused by occlusion and lack of splats, by projecting the erroneous point's current depth to another view, comparing the ground-truth color, and applying foreground splitting or backproject addition. (Sec.~\ref{sec:splat_addition}).}
    \label{fig:pipeline}
\end{figure*}

\subsection{Overview}
Our overall framework is shown in Fig.~\ref{fig:pipeline}. Our method consists of three main components: 4D Gaussian Splatting with shared dynamic transform for a group of splats (Sec.~\ref{sec:dynamic_group}), clustering erroneous pixels to extract effective additional splat candidates (Sec.~\ref{sec:candidate_clustering}), and finally multi-view comparison to add splats by foreground splitting or backproject addition (Sec.~\ref{sec:splat_addition}).

We train a Grouped 4D Gaussian Splatting model, following the progressive learning approach of Ex4DGS~\cite{lee2024ex4dgs}, starting from the first few frames and gradually expanding the model's timespan to encompass the full timespan. Group splitting, 3DGS-like densification, and pruning are done on this stage. In the second step, our pipeline focuses on correcting the errors of the trained Grouped 4D Gaussian Splatting model, with Elliptical Error Clustering and Error-Correcting Splat Addition triggered every few hundred optimization steps. Dynamic and static parameters of groups and splats are still optimized; however, the splat's group membership is frozen in this stage.

\subsection{Grouped 4D Gaussian Splatting}
\label{sec:dynamic_group}

\subsubsection{Grouped Temporal Modeling} We decompose the transform of each splat into two components: a shared dynamic transform at the group level and a relative static transform at the splat level. Specifically, we assign a group-level position and rotation over time, denoted by $\mathbf{x}_{\mathcal{G}}(t)$ and $\mathbf{r}_{\mathcal{G}}(t)$. Each splat has position $\mathbf{x}$, rotation $\mathbf{r}$ relative to the group transform, and a displacement vector $\mathbf{d}$, defined in the global frame.

The group motion is represented using $K$ keyframes: 
$\mathbf{x}_{\mathcal{G}} \in \mathbb{R}^{3 \times K}$ and 
$\mathbf{r}_{\mathcal{G}} \in \mathbb{R}^{4 \times K}$, which are interpolated across time to produce continuous motion. 
The final position and orientation of a splat $g \in \mathcal{G}$ at time $t$ are:
\begin{align}
\mathbf{x}(t) &= \mathbf{x}_{\mathcal{G}}(t) + \mathbf{R}_{\mathcal{G}}(t)\mathbf{x} + t \cdot \mathbf{d}, \\
\mathbf{R}(t) &= \mathbf{R}_{\mathcal{G}}(t) \mathbf{R},
\end{align}
where $\mathbf{R}_{\mathcal{G}}(t)$ and $\mathbf{R}$ are the rotation matrices corresponding to quaternions $\mathbf{r}_{\mathcal{G}}(t)$ and $\mathbf{r}$, respectively.

\subsubsection{Dynamic and Static Splats}
Static splats are implemented by fixing their group transform to the global frame, 
so their relative transform $\mathbf{x}, \mathbf{r}$ is effectively global. 
In this case, the overall parameterization reduces exactly to Ex4DGS’s static formulation with linear motion via $\mathbf{d}$. Dynamic splats additionally have temporal opacity parameters 
$o_c$ (center) and $o_v$ (variance), which control visibility across time.

\subsubsection{Graph-based Dynamic Grouping} We now have to determine which subsets of splats show consistent global motion and should share a group-level transformation. At initialization, all splats are treated as static. As training progresses, we identify splats that move independently from their group. These are detected based on their displacement vectors \(\mathbf{d}\), which capture motion deviating from the shared group trajectory. We train splats with the regularization term similar to Ex4DGS~\cite{lee2024ex4dgs}, splats with consistent movements are discouraged from having large displacements.
Thus, splats exhibiting a large displacement $\mathbf{d}$ are identified as candidates for a new group. 

We construct an undirected graph over these large displacement splats within a group. An edge connects splats $i$ and $j$ when they satisfy two conditions: spatial overlap, and their learned displacements are coherent beyond a threshold $\tau_d$, measured by cosine similarity. Denoting by $\mathbf{x}_i(t)\in\mathbb{R}^3$ the center of splat $i$ at a selected largest mean loss timestamp for group splitting, and by $s(t)$ its effective size, the distance where opacity falls below threshold with Gaussian function along its major axis, the conditions for an edge are: 

\begin{equation} %
\|\mathbf{x}_i(t) - \mathbf{x}_j(t)\|_2 < s_i + s_j
\quad\text{and}\quad
\frac{\mathbf{d}_i^\top \mathbf{d}_j}{\|\mathbf{d}_i\|\|\mathbf{d}_j\|} \;\ge\;\tau_d,
\end{equation}

Upon identification of such connected components, a new dynamic group is formed. The dynamic transform of the split group is initialized by selecting one representative splat from the component. Its relative transform and displacement are merged, and its final transform at keyframes is set as the initial keyframe positions and rotations of the newly formed dynamic group. Subsequently, the inverse of the selected splat's relative transform is applied to the positions and rotations of all splats within this new group to appropriately re-align them relative to the new group's dynamic transform. Furthermore, the chosen splat's displacement is subtracted from the displacements of the splats in the new group, as this motion is now absorbed by the group's shared transform.

\subsection{Elliptical Error Clustering}
\label{sec:candidate_clustering}
Our method improves reconstruction fidelity by first clustering erroneous pixels in the image space into elliptical regions. This stage is critical for accurate and efficient Gaussian splat updates, as elliptical error regions can be directly mapped into additional ellipsoid-shaped Gaussian splats in world space. This section details our clustering strategy, and our splat addition method is introduced in the following section (Sec~\ref{sec:splat_addition}). Our method consists of two steps: i) identifying erroneous pixels using pixel dynamicity and mean RGB error, ii) a recursive clustering and fitting to split them into Gaussian splat candidates.

\subsubsection{Erroneous Pixel Identification} In every few hundred training iterations, we randomly select a viewpoint and compare the rendered result to the ground truth image to identify erroneous regions. These regions are defined using two key criteria: pixel dynamicity and mean RGB error. To focus on dynamic regions, we compute each pixel’s dynamicity as the maximum L1 distance of its ground truth RGB values across the previous, current, and next frames. Pixels exceeding the dynamicity threshold $\tau_D$ are retained. Among these dynamic pixels, we further isolate error-prone areas by applying both an absolute RGB error threshold $\tau_a$ and a relative threshold $\tau_r$, which selects the top percentile of high-error pixels. This filtering process yields a refined set of erroneous pixels suitable for clustering and correction.

\subsubsection{Recursive Pixel Clustering} Filtered pixels are first split into clusters with DBSCAN based on spatial locality. Then, recursive clustering splits them until they are close enough to an ellipse shape. The recursive process consists of: a) pixel clustering by position and color similarity, b) ellipse fitting of clustered pixels. If the ellipse fitting fails, recursion occurs with each cluster split by the first step. This recursion continues until all clusters are a valid ellipse fit.

Initially, we apply spatial clustering using the DBSCAN (Density-Based Spatial Clustering of Applications with Noise) algorithm~\cite{DBSCAN}, which groups pixels based on local density and can discover clusters of arbitrary shape without requiring the number of clusters to be specified in advance. After obtaining the initial set, we further refine it using color information. These initial clusters go through the recursion process below.

First, we combine spatial and color information to group the selected pixels into coherent error clusters. Intuitively, nearby pixels with similar colors are more likely to belong to the same object undergoing consistent motion. We use the DBSCAN for spatial-color clustering, which considers both pixel coordinates and the RGB values (scaled by a constant). This ensures that the final clusters are both spatially coherent and color-consistent, which is essential for fitting elliptical Gaussian splats.

Secondly, for each error cluster obtained from the previous stage, we fit an ellipse to assess its suitability for Gaussian splat correction. Specifically, we compute the minimum-area bounding rectangle of the cluster and evaluate how well the cluster fills an ellipse inscribed within this rectangle. A cluster is accepted as a final elliptical error cluster if its filled area ratio exceeds a predefined threshold, indicating a strong resemblance to an ellipse.

Clusters that fail this test are split into two using K-means clustering on the spatial coordinates. These clusters are recursively fed back into DBSCAN spatial-color clustering and re-evaluated for ellipse fitting to achieve more compact and dense groupings. This iterative process continues until all clusters are accepted. The final set of accepted clusters is parameterized as an ellipse, with their representative color selected from their center pixel for simplicity, as clusters are already color-homogeneous.

\subsection{Error-Correcting Splat Addition}
\label{sec:splat_addition}
This section details our approach to correcting elliptical error clusters by adding or refining existing splats, as depicted in Fig.~\ref{fig:lacking_splat},~\ref{fig:occlusion}. This process targets rendering inaccuracies where the underlying object surface is largely well-reconstructed, but quality suffers from either missing color information or over-reconstructing occlusions.
We take advantage of cross-view comparison to diagnose two primary error types. We sample depths from its center and nearby pixels on the rendered depth map (i.e., alpha-blended depth) for each elliptical cluster, back-projecting them to 3D points. Each 3D point is then projected onto a \textit{comparison view} (from a different camera but the same frame as the \textit{main view} where the error cluster was identified). By comparing the ground-truth pixel colors at the projected 3D location across the main and comparison views, we can distinguish the type of error and devise a targeted fixing strategy.

\subsubsection{Types of Errors} \textit{Lacking Splats (Missing Color)} error occurs when the 3D surface is adequately reconstructed, but correct color information is absent. The defining characteristic is when the ground truth color from the main view largely matches that from the comparison view. This suggests surface consistency across views but inaccurate coloring. \textit{Occlusion (Over-reconstruction)} arises when foreground splats are excessively reconstructed, incorrectly obscuring an object that should be visible. Here, the ground truth color from the main view differs from the comparison view's. This discrepancy implies that the 3D position is not accurately represented, as different background content should be visible from distinct viewpoints.

\subsubsection{Identifying Error Types} To apply the appropriate correction, we extract $k$ depth samples $\{ z_j \}_{j=1}^{k}$ from an $n \times n$ kernel around the ellipse center on the rendered depth map, back-projecting them to 3D points $\{ \mathbf{p}_j \}_{j=1}^{k}$. Each $\mathbf{p}_j$ is then projected to the comparison view for its corresponding ground truth color $\mathbf{c}_\text{comp, j}$. Let $\mathbf{c}_\text{main, j}$ be the ground truth color at the error cluster's ellipse center pixel in the main view.

We apply backproject addition if the following condition, indicating high color similarity across views (thus, a lacking splat error), is met:
\begin{equation}
\min_{j=1,\dots,k} \left\| \mathbf{c}_\text{main, j}- \mathbf{c}_\text{comp, j} \right\|_\infty < \delta_\text{rgb},
\end{equation}
Here, $\|\cdot\|_\infty$ indicates the maximum element-wise difference. Otherwise, foreground splitting is performed, addressing an occlusion error.

\subsubsection{Correcting Error} When the color similarity condition is satisfied, a new splat is initialized to correct the missing color at the 3D point $\mathbf{p}_k$ (corresponding to depth $z_k$) that minimizes $\left\| \mathbf{c}_\text{main, j} - \mathbf{c}_\text{comp, j} \right\|_\infty$. This new splat is attached to the nearest existing splat's group, leveraging the grouped dynamic transform for temporal generalization. Its rotation is initialized to align two axes with the ellipse cluster's projected axes and the other to be perpendicular to it. Its scale is set to create a disk-like shape, covering the ellipse along two axes with a very small scale along the third, ensuring minimal interference with other views at initialization while effectively correcting the error. The splat's opacity is set to $1 - \min_{j=1,\dots,k} \left\| \mathbf{c}_\text{main, j} - \mathbf{c}_\text{comp, j} \right\|_\infty$, allowing other splats to contribute to the remaining difference, and its temporal opacity weight is initialized to be maximal at the selected view's timestamp.

If the color similarity condition is not met, indicating an occlusion error, the splat nearest to the 3D point at the current depth is split. This process is analogous to the densification strategy employed in traditional 3D Gaussian Splatting, effectively refining the problematic region by dividing an existing splat to represent the scene's geometry better and alleviate over-reconstruction.

This addition requires the back-projected 3D point to be visible from the comparison view, reducing the possibility of additional floater overfitting to a specific view. The process is performed based on individual frames and depends on the generalizability of Grouped Temporal Modeling across time. However, each splat has an independent opacity timespan. Splats that fail to generalize across neighboring frames using grouped motion naturally shrink in temporal extent or are split into new groups.

%% file: Sections/05_evaluation.tex
\section{Experiments}
We conducted experiments on the Neural 3D Video and Technicolor dataset, a commonly used benchmark dataset for dynamic novel-view synthesis from multiple cameras. In a similar setup listed in the table, we compared our method with other methods.

We use representative perceptual quality metrics for quantitative evaluation, PSNR, DSSIM, and LPIPS. Some works use a different setup to measure DSSIM, using \textit{scipy}'s function with \textit{data\_range} value $1.0$ and $2.0$ as pointed out in prior works~\cite{attal2023hyperreel, fridovichkeil2023kplanes, li2023spacetime, lee2024ex4dgs}. We show them as \(\text{DSSIM}_1\) and \(\text{DSSIM}_2\) in tables respectively. The model size is for all 50 frames in the Technicolor Light Field dataset and 300 frames in the Neural 3D Video dataset.

\begin{figure*}[h]
    \centering
    \includegraphics[width=1.0\linewidth]{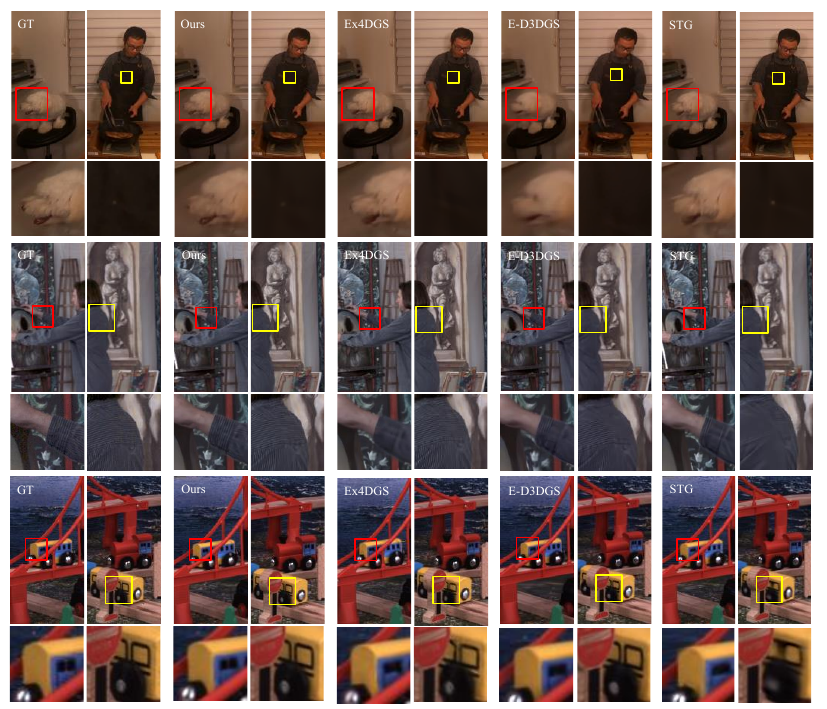}
    \caption{Qualitative comparison of our method with state-of-the-art methods. From the top row, we show examples from the Sear Steak, Painter, and Train scenes from two datasets. The red and yellow box highlights the dynamic area with details, which is shown in a larger size below. For the Painter scene, brightness is adjusted from the ground-truth and rendering result for better visibility. The difference is pronounced in clearer teeth and a button, stripes in clothing, and sharper car window boundaries.}
    \label{fig:qualitative}
\end{figure*}
\subsection{Comparison to State-of-the-arts}
\subsubsection{Technicolor Light Field Dataset}
The Technicolor Light Field dataset contains scenes with dynamic objects covering a large portion of the video, compared to the Neural 3D Video dataset. Our method significantly outperforms previous approaches in this dataset.  Following previous works~\cite{attal2023hyperreel, li2023spacetime}, we use $5$ scenes from the dataset (Painter, Fabien, Theater, Train, Birthday) at an original image resolution of $2024$ by $1088$. The dataset is captured with 4 by 4 camera grid. We select 50 frames from each scene and choose a camera at the second row, second column as a test set, following convention from HyperReel~\cite{attal2023hyperreel}.

Table~\ref{table:technicolo_result} shows the quantitative comparison of our method on the Technicolor Light Field dataset~\cite{Sabater2017technicolor}, with metrics for other methods~\cite{li2022neuraldynerf, attal2023hyperreel, li2023spacetime, lee2024ex4dgs} with metrics from prior works~\cite{lee2024ex4dgs, li2023spacetime} for methods without official results. Our approach significantly improves over the state-of-the-art while being second in terms of DSSIM$_1$. These remarks reflect our method's strong capability to improve quality for highly dynamic and complex objects, such as in the Technicolor dataset.

\begin{table}\small
  \centering
  \caption{Quantitative comparisons on the Technicolor Light Field dataset.}
  \resizebox{0.5\textwidth}{!}{
  \begin{tabular}{@{}lccccc@{}}
  \toprule
  \textbf{Method} & \textbf{PSNR $\uparrow$} & \textbf{DSSIM$_1$ $\downarrow$} & \textbf{DSSIM$_2$ $\downarrow$} & \textbf{LPIPS $\downarrow$} & \textbf{Size $\downarrow$} 
  \\
  \midrule
  DyNeRF~\cite{li2022neuraldynerf}      & 31.80 & \textcolor{gray}{N/A} & 0.021 & 0.140 & \underline{\textbf{30 MB}} 
  \\
  HyperReel~\cite{attal2023hyperreel}   & 32.73 & 0.047 & \textcolor{gray}{N/A} & 0.109 & \underline{60 MB} 
  \\
  4DGS~\cite{yang2023real4dgs}        & 29.54 & 0.065 & 0.032 & 0.149 & \textcolor{gray}{N/A} 
  \\
  4DGaussians~\cite{wu20234d} & 30.79 & 0.079 & 0.040 & 0.178 & \textcolor{gray}{N/A} 
  \\
  STG~\cite{li2023spacetime}         & \underline{33.56} & \textbf{0.040} & \underline{0.019} & \textbf{0.084} & \textbf{55 MB} 
  \\
  SWinGS~\cite{shaw2024swingsslidingwindowsdynamic} & \textbf{33.65} & \underline{\textbf{0.033}} & \textcolor{gray}{N/A} & 0.117 & \textcolor{gray}{N/A} 
  \\
  E-D3DGS~\cite{bae2024ed3dgs}     & 33.24 & 0.047 & \textcolor{gray}{N/A} & 0.100 & 77 MB 
  \\
  Ex4DGS~\cite{lee2024ex4dgs}      & 33.62 & 0.042 & \underline{0.019} & \underline{0.088} & 144 MB 
  \\
  Ours        & \underline{\textbf{34.04}} & \textbf{0.040} & \underline{\textbf{0.018}} & \underline{\textbf{0.081}} & 177 MB 
  \\
  \bottomrule
  \end{tabular}
  }
  \label{table:technicolo_result}
\end{table}

\subsubsection{Neural 3D Dataset}
Neural 3D Video dataset~\cite{neural3dvideo} is a widely used dataset for multi-view novel view synthesis with 18 to 21 cameras. The dataset contains video of original resolution of 2704 by 2028, and downsampled to a half resolution as convention. We follow the recent convention~\cite{lee2024ex4dgs} for processing the dataset. Some prior works~\cite{neuralvolumes, llff, li2022neuraldynerf, cao2023hexplane, im4d, 4k4d} show the evaluation results with different sets of scenes. Thus, we also show metrics for a commonly used subset of the dataset. The scene set used to compute the metric is denoted with a superscript for those evaluations. Per-scene metric is available in the supplementary material.

Table~\ref{tab:neural3d_comparison} compares our method to various approaches. Despite a small dynamic area to improve, our method achieves the best PSNR and outperforms baseline Ex4DGS~\cite{lee2024ex4dgs} in visual quality, while keeping the benefit from using sparse COLMAP point cloud input from only one frame, minimizing the pre-processing time and burden of per-frame COLMAP reconstruction size. Our method is also comparable to the state-of-the-art techniques in other metrics. The file size remains similar to the baseline because of reduced dynamic parameters despite additional splats.
\begin{table}[ht]
\centering
\caption{Quantitative comparisons on the Neural 3D Video Dataset.\textsuperscript{1}Only the \textit{Flame Salmon} scene. \textsuperscript{2}Except for the \textit{Coffee Martini} scene. \textsuperscript{3}Only the \textit{Cut Roasted Beef} scene.}
\label{tab:neural3d_comparison}
\resizebox{0.5\textwidth}{!}{
\begin{tabular}{lcccccc}
\toprule
\textbf{Method} & \textbf{PSNR $\uparrow$} & \textbf{DSSIM$_1$ $\downarrow$} & \textbf{DSSIM$_2$ $\downarrow$} & \textbf{LPIPS $\downarrow$} & \textbf{Size $\downarrow$}
\\
\midrule
Neural Volumes\textsuperscript{1}~\cite{neuralvolumes} & 22.80 & \textcolor{gray}{N/A} & 0.062 & 0.295 & \textcolor{gray}{N/A}
\\
LLFF\textsuperscript{1}~\cite{llff}           & 23.24 & 0.076 & \textcolor{gray}{N/A} & 0.235 & \textcolor{gray}{N/A}
\\
DyNeRF\textsuperscript{1}~\cite{li2022neuraldynerf}         & 29.58 & \textcolor{gray}{N/A} & 0.020 & 0.083 & 28 MB
\\
Ours\textsuperscript{1} & 29.49 & 0.040 & 0.022 & 0.067 & 125 MB
\\
\midrule
HexPlane\textsuperscript{2}~\cite{cao2023hexplane}       & 31.71 & \textcolor{gray}{N/A} & \textcolor{gray}{N/A} & 0.075 & 200 MB
\\
Ours\textsuperscript{2} & 32.86 & 0.025 & 0.013 & 0.044 & 114 MB
\\
\midrule
Im4D\textsuperscript{3}~\cite{im4d}           & 32.58 & \textcolor{gray}{N/A} & \textcolor{gray}{N/A} & \textcolor{gray}{N/A} & \textcolor{gray}{N/A}
\\
4K4D\textsuperscript{3}~\cite{4k4d}           & 32.86 & \textcolor{gray}{N/A} & \textcolor{gray}{N/A} & \textcolor{gray}{N/A} & \textcolor{gray}{N/A}
\\
Ours\textsuperscript{3} & 33.78 & 0.022 & 0.011 & 0.040 & 119 MB
\\
\midrule
NeRFPlayer~\cite{nerfplayer}                        & 30.69 & 0.034 & \textcolor{gray}{N/A}     & 0.105 & 5130 MB
\\
HyperReel~\cite{attal2023hyperreel}                         & 31.10 & 0.036 & \textcolor{gray}{N/A}     & 0.096 & 360 MB
\\
K-Planes~\cite{fridovichkeil2023kplanes}                          & 31.63 & 0.041 & \textcolor{gray}{N/A}    & 0.110 & 311 MB
\\
MixVoxels-L~\cite{mixvoxels}                        & 31.34 & 0.036 & \textcolor{gray}{N/A}     & 0.087 & 500 MB
\\
MixVoxels-X~\cite{mixvoxels}                       & 31.73 & 0.032 & \textcolor{gray}{N/A}     & 0.078 & 500 MB
\\
4DGS~\cite{yang2023real4dgs}            & 32.01 & \textcolor{gray}{N/A} & \textbf{\underline{0.014}} & 0.055 & 6270 MB
\\
4DGaussians~\cite{wu20234d}     & 31.15  & \textcolor{gray}{N/A} & 0.016 & 0.049 & 34 MB
\\
Rotor4DGS~\cite{duan:2024:4drotorgs} & 31.62 & 0.030 & \textcolor{gray}{N/A} & 0.140 & \textcolor{gray}{N/A}
\\
STG~\cite{li2023spacetime}                              & \underline{32.05} & \textbf{\underline{0.026}} & \textbf{\underline{0.014}} & \textbf{0.044} & 107 MB
\\
SWinGS~\cite{shaw2024swingsslidingwindowsdynamic}                             & 31.10 & 0.030  & \textcolor{gray}{N/A} & 0.096 & \textcolor{gray}{N/A}
\\
E-D3DGS~\cite{bae2024ed3dgs} & 31.31 & \textbf{0.028}  & \textcolor{gray}{N/A} & \textbf{\underline{0.037}} & 66 MB
\\
Ex4DGS~\cite{lee2024ex4dgs} & \textbf{32.11} & 0.030 & \underline{0.015} & 0.048 & 115 MB
\\
Ours & \textbf{\underline{32.23}} & \textbf{0.028} & \underline{0.015} & \underline{0.047} & 115 MB
\\
\bottomrule
\end{tabular}
}
\end{table}

\subsubsection{Qualitative Results}
We qualitatively compare our method with state-of-the-art methods with publicly available models, E-D3DGS~\cite{bae2024ed3dgs}, Ex4DGS~\cite{lee2024ex4dgs}, and the STG~\cite{li2023spacetime} model trained with official code, since models are not available for most scenes. Fig.~\ref{fig:qualitative} and the supplementary video provide qualitative results of our method. Qualitative results show \methodname's capability to reconstruct dynamic parts in the scene in detail.

\subsection{Analysis}
\label{sec:analysis}

\begin{table}\small
  \caption{Ablation study of our method regarding its visual quality and efficiency of splat addition. We conducted experiments with the Technicolor Light Field~\cite{Sabater2017technicolor} dataset.}
  \centering
\resizebox{0.5\textwidth}{!}{
  \begin{tabular}{@{}lccccc@{}}
  \toprule
  \textbf{Method} & \textbf{PSNR↑} & \textbf{DSSIM$_1$↓} & \textbf{LPIPS↓} & \textbf{N Splats↓} \\
  \midrule
  Baseline & 33.62 & 0.0418 & 0.0878 & \textbf{426K} \\
  Baseline Lower Threshold & 33.61 & 0.0410 & 0.0845 & 593K \\
  \midrule
  Without Group & 33.81 & 0.0407 & 0.0840 & 502K \\
  \midrule
  Pixel-wise Backprojection & 33.76 & 0.0409 & 0.0807 & 600K \\
  Patch-wise Backprojection & 33.78 & 0.0411 & 0.0820 & 551K \\
  Without Color-aware Clustering & 33.81 & 0.0408 & 0.0813 & 561K \\
  Without Shape-based Division & 33.93 & 0.0404 & 0.0813 & 558K \\
  \midrule
  Only Backproject Addition & 33.89 & 0.0404 & \textbf{0.0804} & 561K \\
  Only Foreground Splitting & 33.84 & 0.0408 & 0.0821 & 551K \\
  \midrule
  Without Error-based Correction & 33.66 & 0.0415 & 0.0860 & 550K \\
  \midrule
  Ours & \textbf{34.04} & \textbf{0.0401} & 0.0809 & 565K \\
  \bottomrule
  \end{tabular}
  \label{table:ablation}
  }
\end{table}

\label{sec:ablation}
\subsubsection{Ablation Studies}
We conduct experiments to show the impact of components in pipelines, with metrics summarized in Table~\ref {table:ablation}. To show the effectiveness of our error-based correction, we summarize the perpetual quality metrics and a number of splats.

First, we show the baseline method~\cite{lee2024ex4dgs} performance, with original and a lower dynamic splat densification threshold. Vanilla densification with more splat budget, \textit{Baseline Lower Threshold}, does not identify appropriate spots to add splats. The shows a clear advantage of our specialized method of adding splats to gain quality improvement from additional splats. Interestingly, using a lower densification threshold improved DSSIM and LPIPS, while PSNR remained relatively unchanged, indicating a high pixel error, despite being supervised to minimize it. However, the quality improvement does not match our approach, despite having more splats.

We further experimented with the importance of our groupings introduced in Sec.~\ref{sec:dynamic_group}, keeping the error correction method while ablating the grouped dynamic representation. \textit{Without Grouping}, our error correction method is significantly affected. Combined with this result, we demonstrate the impact of grouping on improving consistent splat-object correspondence in Sec.~\ref{sec:consistency}.

We experiment to show the importance of Elliptical Error Clustering in Sec.~\ref{sec:candidate_clustering}. \textit{Pixel-wise Backprojection} uses our algorithm without clustering, creating error correction candidates per pixel instead. It results in significantly more splats. \textit{Patch-wise Backprojection} uses ellipse fitting per error points within a patch of size 16 by 16, similar to the error sampling method by STG~\cite{li2023spacetime}. Patch-wise projection adds similar splats to our method. Neither approach yielded a significant improvement. \textit{Without Color-aware Clustering} creates clusters only based on the proximity, with an initial DBSCAN pass and K-Means splitting to fit an elliptical shape, which can result in non-uniform colored error within the cluster. \textit{Without Shape-based Division} does not split clusters based on their similarity to an ellipse, resulting in shapes incompatible with splats. These experiments show that our clustering tailored for splat addition is crucial to pinpoint where splat should be added.

Finally, we analyze the impact of our Error-Correcting Splat Addition approach from Sec.~\ref{sec:splat_addition}. Generally, combining the two approaches performs better than using one for the rest of the scenes. However, the importance of the two approaches varied. Interestingly, \textit{Only Backproject Addition} had better LPIPS on average than the full method. For the Train scene, using \textit{Only Foreground Splitting} had a similar result to the full method. LPIPS is better than the full method for the Birthday scene, but has worse PSNR.
\textit{Without Error-based Correction}, using only grouped dynamic transform, a slight improvement is observed compared to the baseline, resulting in more splats. With improved correspondence of splats with dynamic objects, dynamic object representation could be better; however, reduced dynamic freedom requires more splats.

\begin{figure*}[ht]
    \centering
    \includegraphics[width=1.0\linewidth]{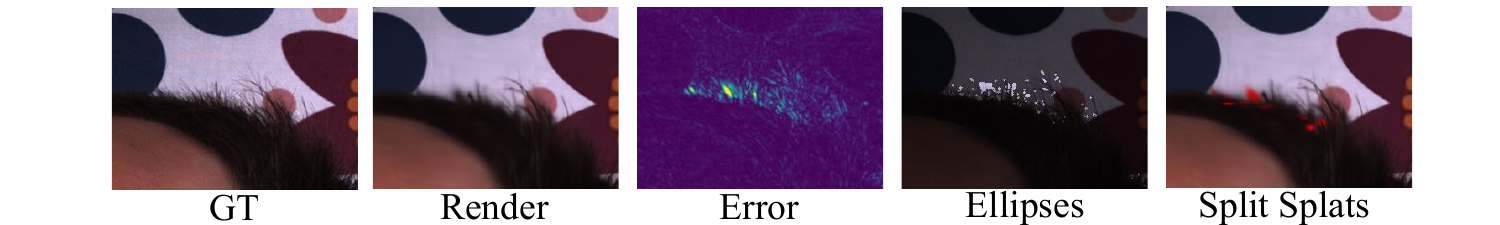}
    \caption{Example of Foreground Splitting with our method in the Fabien scene. Hair has inconsistent details with the ground truth, occluding the background. Based on the error, ellipses are clustered for the correction candidate. Splats for hair are split (rendered with red color in the image).}
    \label{fig:split_visualize}
\end{figure*}

\begin{figure*}[ht]
    \centering
    \includegraphics[width=0.75\linewidth]{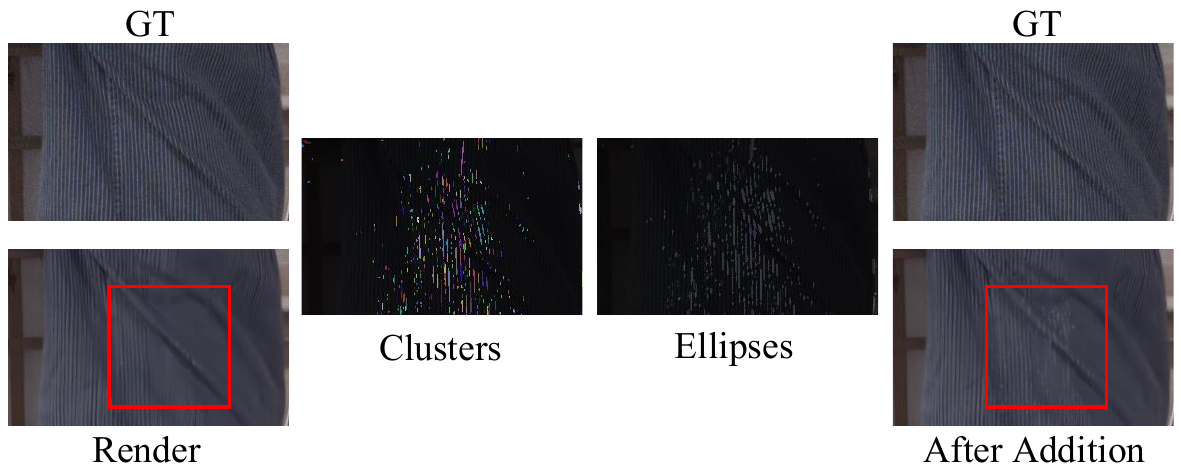}
    \caption{Example of Backproject Addition with our method in the Painter scene with brightness adjusted for visibility. The rendering result shows stripes missing on the clothes. Our method clusters those errors and generates ellipses. Finally, after applying our splat addition method, the dynamic object has a corrected surface. This can be further optimized to reduce errors.}
    \label{fig:addition_visualize}
\end{figure*}
\subsubsection{Visualization of Clustering and Splat Addition}
\label{sec:splat_add_viz}
We visualize the error correction process with figures. In the Foreground Splitting case (Fig.~\ref{fig:split_visualize}), hair is inaccurately represented, occluding the background. Checking the color consistency of the center pixel of the ellipse error cluster with other views, if the color appears inconsistent, the splat in the foreground, for hair, is split. The split splats are marked with red color. Further optimization with additional splats can provide a more accurate fit for the hair. Not all candidates are split due to a visibility constraint from the comparison view; multiple attempts from different views gradually improve details. For the Backproject Addition case (Fig.~\ref{fig:addition_visualize}), the Grouped 4DGS exhibits a lack of splat on the surface of clothes, lacking stripes. The elliptical error clustering approach identifies erroneous areas and divides them into ellipses. The stripe is visible after adding splats based on the Elliptical Error Clusters. Repeating the error correction process and optimization, our method gradually corrects the dynamic object from the most erroneous area.

\begin{figure*}[ht]
    \centering
    \includegraphics[width=0.8\linewidth]{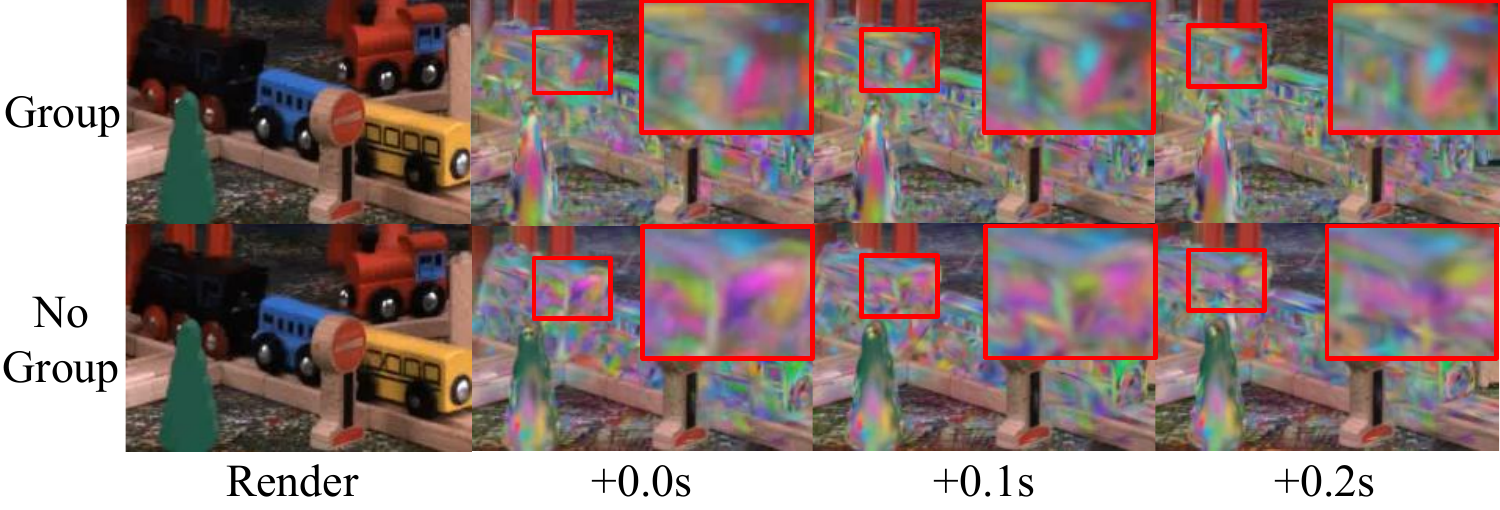}
    \includegraphics[width=0.8\linewidth]{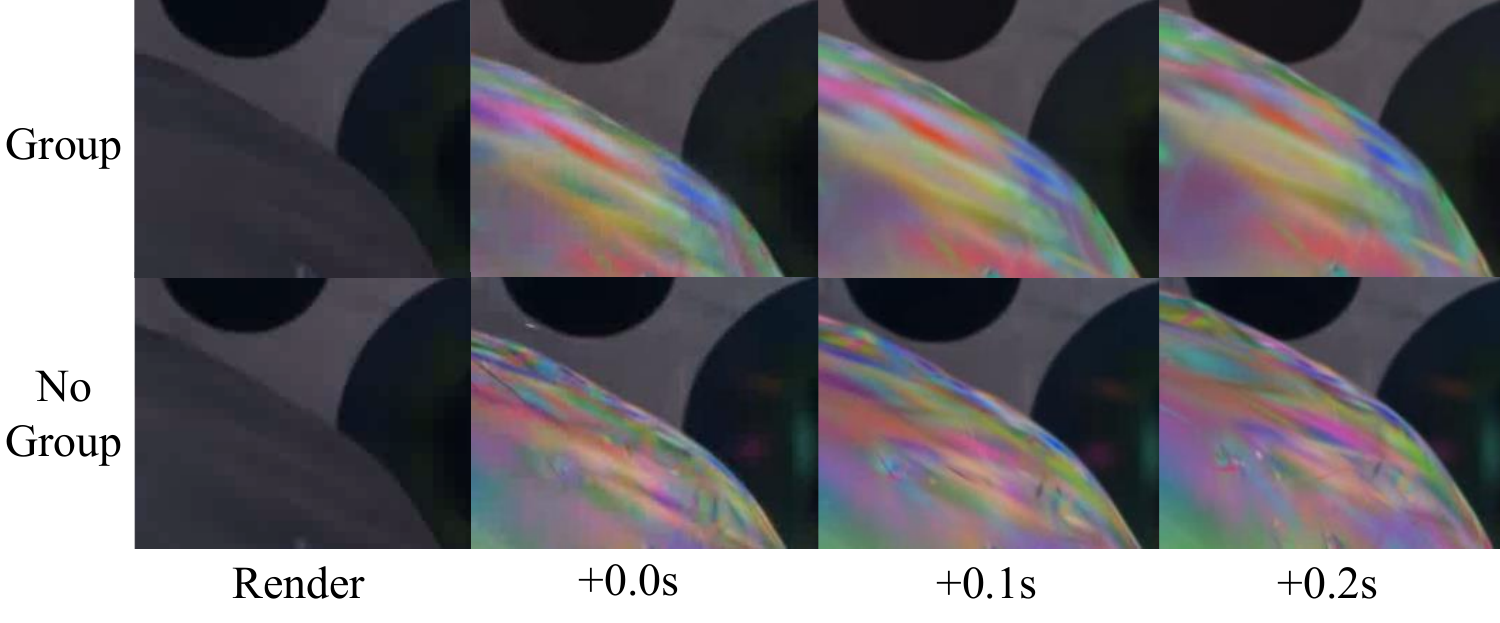}
    \caption{Comparison of object-splat correspondence consistency of our group-based representation and baseline~\cite{lee2024ex4dgs}, with 3 frame gaps (0.1 sec). In the first example in the Train scene, the red box focuses on a small area of the railroad car, featuring a mostly uniform color and small details. Comparison of splat alignment in different frames reveals that the grouped transform enables splats to follow the real object more faithfully. For the second example in Fabien's scene, clothes are represented with mostly the same splats with grouping, while the baseline exhibits consistently changing splats, particularly near the boundary area.}
    \label{fig:qualitative_consistency}
\end{figure*}
\subsubsection{Splat-Object Correspondence Consistency} \label{sec:consistency} We demonstrate that the Grouped 4D Gaussian Splatting continuously tracks dynamic components in the scene by grouping the dynamic transform. Fig.~\ref{fig:qualitative_consistency} visualizes the splat and the correspondence between the displayed objects. For visualization, scenes are rendered with a random, unique color assigned for each dynamic splat to visualize how they move in the scene. Train and Fabien scene from the Technicolor Light Field dataset are visualized every three frames (0.1 sec).

Our method consistently assigns a splat to the same object across different frames with our grouped dynamic representation. The baseline per-splat method shows temporal jitter in this visualization. Multiple splats with the wrong dynamic transform and new splats appearing and disappearing in the area to represent one dynamic object indicate a failure to follow the dynamic transform of the real-world object correctly.

\subsubsection{Runtime Measurements and Temporal Stability}
\begin{table}[t]
\centering
\caption{Comparison of training, rendering, memory, and temporal stability.}
\label{tab:more_metric_technicolor}
\begin{tabular}{lcc}
\toprule
\textbf{Metric} & \textbf{Baseline~\cite{lee2024ex4dgs}} & \textbf{Ours} \\
\midrule
Training (h) & 1.90 & 3.21 \\
Rendering (s/frame) & 0.0140 & 0.0210 \\
Memory (GB) & 2.98 & 3.37 \\
Temporal Stability (tPSNR) & 37.43 & 37.60 \\
\bottomrule
\end{tabular}
\end{table}

We measured training time, rendering time, allocated memory during training using a NVIDIA RTX A6000 GPU for measurements, and tPSNR metric~\cite{tpsnr} for the Technicolor Light Field dataset for comparison to the baseline in Table~\ref{tab:more_metric_technicolor}. The tPSNR metric is measured by computing PSNR with the ground-truth and the rendered image's residual between consecutive frames:
\begin{equation}
PSNR(\text{render}_{t}-\text{render}_{t-1}, \text{gt}_{t}-\text{gt}_{t-1})
\end{equation}

We trained for $10,000$ additional epochs for the Error Correction stage in addition to the original epochs for the baseline. Additional epochs and splats are reflected in the proportional relevant time and memory. Following grouped motion may come with temporal instability when incorrect splats are added to groups. Group splitting and temporal opacity weight for individual splats effectively address this issue, resulting in improved temporal stability compared to the baseline.

%% file: Sections/06_conclusion.tex
\section{Conclusion and Limitations}
We have presented a novel approach to enhance dynamic novel-view synthesis with 4D Gaussian Splatting by explicitly addressing the critical limitations of ambiguous temporal correspondences and ineffective splat densification. Our method introduces elliptical clustering to localize regions requiring additional splats precisely and employs group-based temporal modeling to enforce stable, coherent splat correspondences across frames. Through targeted error corrections—backprojection-based addition for missing-color errors and foreground splitting for occlusion errors—our method substantially improves perceptual rendering quality and temporal consistency. Experimental evaluations on widely used benchmarks demonstrate significant improvements, up to 0.39dB of PSNR over state-of-the-art methods, particularly in reducing visual artifacts and temporal jitter. We believe our contributions open promising directions for future research, including further optimization of splat efficiency and enhanced handling of complex dynamic scenes.

Our methods have limitations. The proposed formulation is best suited for rigid or geometrically contiguous deformations. Our method assumes a fixed color per splat and keyframed motion, following the baseline approach~\cite{lee2024ex4dgs}, which limits the ability to model significant appearance changes. As shared by existing 3DGS and 4DGS approaches, translucent objects and volumetric effects (e.g., flames) remain challenging.
Yet, our error-driven splat addition with group-level motion modeling provides a flexible foundation that can be adapted to more advanced representations, such as those with learned appearance or volumetric support.

%% file: Sections/99_supplementary.tex
\appendix
\maketitle
\section{Sections}
This supplementary material provides the following:
\begin{itemize}
\item{Dynamic Splat Property Formulation: Introducing formulation of how the baseline method and our method handle temporally varying properties.}
\item{Per Scene Evaluation Results: We provide PSNR, SSIM, and LPIPS evaluation results for each scene.}
\end{itemize}

We provide video files as supplementary material. The video files consist of visualizations of the consistency of dynamic object-splat correspondence, comparing Grouped 4D Gaussian Splatting and the baseline. It also shows a video version of the qualitative comparison of our method, particularly in a dynamic part of the video.

\section{Dynamic Property Formulation}
We adopt Ex4DGS~\cite{lee2024ex4dgs}'s parameterization to represent dynamic transforms of splats, and we present the formulation of time-variant properties here.

The dynamic splat group's properties, position, and rotation are interpolated across time; $K$ keyframes are provided every $I$ frames. Temporal opacity is modeled continuously as a function of time \(t \in [0, T]\) per splat.

\subsection{Position Interpolation}
Given keyframed positions:
\begin{equation}
\mathbf{x}_{\mathcal{G},0}, \mathbf{x}_{\mathcal{G},1}, \ldots, \mathbf{x}_{\mathcal{G},K-1} \in \mathbb{R}^3
\end{equation}
The dynamic position \(x_d(t)\) is interpolated using cubic Hermite splines, which provide smooth transitions using both values and tangents at keyframes. For a given time \(t\), keyframe index for reference is computed:
\[
n = \left\lfloor \frac{t}{I} \right\rfloor, \quad t' = \frac{t - nI}{I}, \quad t' \in [0, 1]
\]
Denoting the cubic Hermite interpolator as $\mathcal{H}$, The interpolated position is:
\begin{align}
\mu(t) = \mathcal{H}(\mathbf{x}_{\mathcal{G},n}, \boldsymbol{\delta}_{\mathcal{G},n}, \mathbf{x}_{\mathcal{G},n+1}, \boldsymbol{\delta}_{\mathcal{G},n+1};\ t')\end{align}
with tangent vectors estimated with finite differences:
\begin{align}
\boldsymbol{\delta}_{\mathcal{G},n} = \frac{\mathbf{x}_{\mathcal{G},n+1} - \mathbf{x}_{\mathcal{G},n-1}}{2I}\end{align}
The cubic Hermite interpolation function is defined as:
\begin{align}
\mathcal{H}(\mathbf{x}_0, \boldsymbol{\delta}_0, \mathbf{x}_1, \boldsymbol{\delta}_1;\ t) =\ 
& (2t^3 - 3t^2 + 1)\mathbf{x}_0 + (t^3 - 2t^2 + t)\boldsymbol{\delta}_0 \\
& + (-2t^3 + 3t^2)\mathbf{x}_1 + (t^3 - t^2)\boldsymbol{\delta}_1
\end{align}

\subsection{Rotation Interpolation}
Given keyframed quaternions:
\begin{equation}
    \mathbf{r}_{\mathcal{G},0}, \mathbf{r}_{\mathcal{G},1}, \ldots, \mathbf{r}_{\mathcal{G},K-1} \in \mathbb{R}^4
\end{equation}
Intermediate rotations are interpolated using spherical linear interpolation (Slerp)~\cite{slerp}. For a given time \(t\), the index of the keyframe to refer to is computed:
\begin{align}
n = \left\lfloor \frac{t}{I} \right\rfloor, \quad t' = \frac{t - nI}{I}
\end{align}
Let \(\mathbf{q}_0 = \mathbf{r}_{\mathcal{G},n}\) and \(\mathbf{q}_1 = \mathbf{r}_{\mathcal{G},n+1}\). Then, the interpolated quaternion is:
\begin{align}
\mathbf{q}_0' = \frac{\mathbf{q}_0}{\|\mathbf{q}_0\|}, \quad
\mathbf{q}_1' = \frac{\mathbf{q}_1}{\|\mathbf{q}_1\|}, \quad
\theta = \cos^{-1}(\langle \mathbf{q}_0', \mathbf{q}_1' \rangle) \\
\mathbf{q}(t) = \frac{\sin((1 - t')\theta)}{\sin\theta} \mathbf{q}_0' + \frac{\sin(t' \theta)}{\sin\theta} \mathbf{q}_1'
\end{align}

where $\mathbf{q}_0'$ and $\mathbf{q}_1'$ is normalized keyframe quaternion and \(\theta\) is the angle between the two quaternions.

\subsection{Temporal Opacity}
Time-dependent opacity is modeled as the product of a base opacity \(o\) and a continuous temporal weighting function \(o_w(t)\). The function \(o_w(t)\) smoothly modulates the splat's visibility using two Gaussian falloffs centered at the opacity centers \(o_{c, 0}\) and \(o_{c, 1}\), which represent the temporal bounds of the splat's full opacity interval. The parameters \(o_{v, 0}\) and \(o_{v, 1}\) are the variances controlling the smoothness of the fade-in before \(o_{c, 0}\) and fade-out after \(o_{c, 1}\), respectively. The opacity weight function is defined as:
\begin{align}
o(t) &= o \cdot o_w(t) \\
o_w(t) &=
\begin{cases}
\exp\left(-\frac{(t - o_{c, 0})^2}{o_{v, 0}^2}\right), & t < o_{c, 0} \\
1, & o_{c, 0} \leq t \leq o_{c, 1} \\
\exp\left(-\frac{(t - o_{c, 1})^2}{o_{v, 1}^2}\right), & t > o_{c, 1}
\end{cases}
\end{align}
This formulation ensures smooth and differentiable visibility control over time by fading in and out the splat opacity outside the temporal interval \([o_{c, 0}, o_{c, 1}]\).

\section{Per Scene Evaluation Results}
For SSIM, we use a metric of scipy SSIM with $range=1$ shared by recent papers~\cite{li2023spacetime, bae2024ed3dgs, lee2024ex4dgs}. We present the state-of-the-art method showing per-scene respective metric in the paper. The best metric for each scene is marked in bold.

\begin{table*}[ht]
\centering
\caption{Per-scene PSNR comparison on Technicolor Light Field dataset (higher is better).}
\label{tab:psnr}
\begin{tabular}{lccccc|c}
\toprule
\textbf{Model} & \textbf{Birthday} & \textbf{Fabien} & \textbf{Painter} & \textbf{Theater} & \textbf{Train} & \textbf{Average} \\
\midrule
DyNeRF~\cite{li2022neuraldynerf} & 29.20 & 32.76 & 35.95 & 29.53 & 31.58 & 31.80 \\
HyperReel~\cite{attal2023hyperreel} & 29.99 & 34.70 & 35.91 & \textbf{33.32} & 29.74 & 32.73 \\
STG~\cite{li2023spacetime} & 32.09 & \textbf{35.70} & 36.44 & 30.99 & \textbf{32.58} & 33.60 \\
4DGS~\cite{wu20234d} & 28.01 & 26.19 & 33.91 & 31.62 & 27.96 & 29.54 \\
4DGaussians~\cite{yang2023real4dgs} & 30.87 & 33.56 & 34.36 & 29.81 & 25.35 & 30.79 \\
E-D3DGS~\cite{bae2024ed3dgs} & \textbf{32.90} & 34.71 & 36.18 & 31.07 & 31.33 & 33.23 \\
Ex4DGS~\cite{lee2024ex4dgs} & 32.38 & 35.38 & 36.73 & 31.84 & 31.77 & 33.62\\
\midrule
\textbf{Ours} & 32.83 & 35.67 & \textbf{37.08} & 32.28 & 32.32 & \textbf{34.04} \\
\bottomrule
\end{tabular}
\end{table*}

\begin{table*}[ht]
\centering
\caption{Per-scene SSIM$_1$ comparison on Technicolor Light Field dataset (lower is better).}
\label{tab:ssim1}
\begin{tabular}{lccccc|c}
\toprule
\textbf{Model} & \textbf{Birthday} & \textbf{Fabien} & \textbf{Painter} & \textbf{Theater} & \textbf{Train} & \textbf{Average} \\
\midrule
HyperReel~\cite{attal2023hyperreel} & 0.922 & 0.895 & 0.923 & \textbf{0.895} & 0.895 & 0.906 \\
STG~\cite{li2023spacetime} & 0.942 & \textbf{0.906} & 0.927 & 0.881 & 0.941 & 0.919 \\
4DGS~\cite{wu20234d} & 0.902 & 0.856 & 0.897 & 0.869 & 0.843 & 0.873 \\
4DGaussians~\cite{yang2023real4dgs} & 0.904 & 0.854 & 0.884 & 0.841 & 0.730 & 0.843 \\
E-D3DGS~\cite{bae2024ed3dgs} & \textbf{0.951} & 0.885 & 0.924 & 0.868 & 0.912 & 0.907 \\
Ex4DGS~\cite{lee2024ex4dgs} & 0.943 & 0.889 & 0.929 & 0.880 & 0.937 & 0.916 \\
\midrule
\textbf{Ours} & 0.948 & 0.892 & \textbf{0.932} & 0.886 & \textbf{0.942} & \textbf{0.920} \\
\bottomrule
\end{tabular}
\end{table*}

\begin{table*}[ht]
\centering
\caption{Per-scene LPIPS comparison on Technicolor Light Field dataset (lower is better).}
\label{tab:lpips}
\begin{tabular}{lccccc|c}
\toprule
\textbf{Model} & \textbf{Birthday} & \textbf{Fabien} & \textbf{Painter} & \textbf{Theater} & \textbf{Train} & \textbf{Average} \\
\midrule
DyNeRF~\cite{li2022neuraldynerf} & 0.067 & 0.242 & 0.146 & 0.188 & 0.067 & 0.142 \\
HyperReel~\cite{attal2023hyperreel} & 0.053 & 0.186 & 0.117 & \textbf{0.115} & 0.072 & 0.109 \\
STG~\cite{li2023spacetime} & 0.042 & 0.114 & 0.096 & 0.133 & \textbf{0.037} & 0.084 \\
4DGS~\cite{wu20234d} & 0.089 & 0.197 & 0.136 & 0.155 & 0.166 & 0.149 \\
4DGaussians~\cite{yang2023real4dgs} & 0.087 & 0.186 & 0.161 & 0.187 & 0.271 & 0.178 \\
Ex4DGS~\cite{lee2024ex4dgs} & 0.044 & 0.123 & 0.091 & 0.129 & 0.052 & 0.088 \\
\midrule
\textbf{Ours} & \textbf{0.038} & \textbf{0.120} & \textbf{0.089} & 0.116 & 0.043 & \textbf{0.081} \\
\bottomrule
\end{tabular}
\end{table*}

\begin{table*}[ht]
\centering
\caption{Per-scene PSNR comparison on the Neural 3D Video dataset (higher is better).}
\label{tab:per_scene_psnr}
\resizebox{\textwidth}{!}{
\begin{tabular}{lcccccc|c}
\toprule
\textbf{Model} & \textbf{Coffee Martini} & \textbf{Cook Spinach} & \textbf{Cut Roasted Beef} & \textbf{Flame Salmon} & \textbf{Flame Steak} & \textbf{Sear Steak} & \textbf{Average} \\
\midrule
NeRFPlayer~\cite{nerfplayer} & \textbf{31.53} & 30.56 & 29.35 & \textbf{31.65} & 31.93 & 29.13 & 30.69 \\
STG~\cite{li2023spacetime} & 28.61 & 32.92 & 33.52 & 29.48 & 33.64 & 33.89 & 32.05 \\
4DGS~\cite{wu20234d} & 28.33 & 32.93 & \textbf{33.85} & 29.38 & \textbf{34.03} & 33.51 & 32.01 \\
4DGaussians~\cite{yang2023real4dgs} & 27.34 & 32.46 & 32.90 & 29.20 & 32.51 & 32.49 & 31.15 \\
E-D3DGS~\cite{bae2024ed3dgs}& 29.10 & 32.96 & 33.57 & 29.61 & 33.57 & 33.45 & 31.31 \\
Ex4DGS~\cite{lee2024ex4dgs} & 28.79 & \textbf{33.23} & 33.73 & 29.29 & 33.91 & 33.69 & 32.11 \\
\midrule
Ours            & 29.07 & 33.05 & 33.78 & 29.49 & 33.89 & \textbf{34.10} & \textbf{32.23} \\
\bottomrule
\end{tabular}
}
\end{table*}

\begin{table*}[t]
\centering
\caption{Per-scene SSIM$_1$ comparison on Neural 3D Video dataset (higher is better).}
\resizebox{\textwidth}{!}{
\label{tab:ssim_comparison}
\begin{tabular}{lcccccc|c}
\toprule
\textbf{Method} & \textbf{Coffee Martini} & \textbf{Cook Spinach} & \textbf{Cut Roasted Beef} & \textbf{Flame Salmon} & \textbf{Flame Steak} & \textbf{Sear Steak} & \textbf{Average} \\
\midrule
STG~\cite{li2023spacetime} & 0.917 & \textbf{0.957} & \textbf{0.959} & 0.925 & \textbf{0.965} & \textbf{0.966} & \textbf{0.948} \\
E-D3DGS~\cite{bae2024ed3dgs} & \textbf{0.931} & 0.956 & 0.958 & \textbf{0.936} & 0.964 & 0.963 & 0.945 \\
4DGaussians~\cite{yang2023real4dgs} & 0.905 & 0.949 & 0.957 & 0.917 & 0.954 & 0.957 & 0.940 \\
Ex4DGS~\cite{lee2024ex4dgs} & 0.915 & 0.947 & 0.948 & 0.917 & 0.956 & 0.959 & 0.940 \\
\midrule
Ours            & 0.913 & 0.953 & 0.956 & 0.919 & 0.960 & 0.958 & 0.944 \\
\bottomrule
\end{tabular}
}
\end{table*}

\begin{table*}[t]
\centering
\caption{Per-scene LPIPS comparison on Neural 3D Video dataset (lower is better).}
\resizebox{\textwidth}{!}{
\label{tab:lpips_comparison}
\begin{tabular}{lcccccc|c}
\toprule
\textbf{Method} & \textbf{Coffee Martini} & \textbf{Cook Spinach} & \textbf{Cut Roasted Beef} & \textbf{Flame Salmon} & \textbf{Flame Steak} & \textbf{Sear Steak} & \textbf{Average} \\
\midrule
NeRFPlayer~\cite{nerfplayer} & 0.085 & 0.113 & 0.144 & 0.098 & 0.088 & 0.138 & 0.111 \\
MixVoxels-L~\cite{mixvoxels} & 0.106 & 0.099 & 0.088 & 0.116 & 0.088 & 0.080 & 0.096 \\
4DGS~\cite{wu20234d} & 0.079 & 0.041 & 0.041 & 0.078 & 0.036 & 0.037 & 0.052 \\
STG~\cite{li2023spacetime} & \textbf{0.069} & \textbf{0.037} & \textbf{0.036} & \textbf{0.063} & \textbf{0.029} & \textbf{0.030} & \textbf{0.044} \\
4DGaussians~\cite{yang2023real4dgs} & 0.095 & 0.056 & 0.104 & 0.095 & 0.050 & 0.046 & 0.074 \\
Ex4DGS~\cite{lee2024ex4dgs} & 0.070 & 0.042 & 0.040 & 0.066 & 0.034 & 0.035 & 0.048 \\
\midrule
Ours            & \textbf{0.069} & 0.042 & 0.040 & 0.067 & 0.033 & 0.035 & 0.047 \\
\bottomrule
\end{tabular}
}
\end{table*}